\documentclass[conference]{IEEEtran}
\IEEEoverridecommandlockouts
\usepackage{amsmath,amsfonts}
\usepackage{array}
\usepackage{textcomp}
\usepackage{stfloats}
\usepackage{url}
\usepackage{graphicx}
\usepackage[normalem]{ulem}
\useunder{\uline}{\ul}{}
\usepackage{verbatim}
\usepackage{graphicx}
\usepackage{amssymb}
\usepackage{hyperref}
\usepackage{graphicx}
\usepackage{amsmath}
\usepackage{longtable}
\usepackage{algorithm} 
\usepackage{algpseudocode} 
\usepackage{mathrsfs}
\usepackage{subcaption}
\usepackage{mathtools}
\usepackage{pifont}
\usepackage{color}
\usepackage{lineno}
\usepackage{makecell} 
\usepackage{pdflscape}
\usepackage{adjustbox}
\usepackage[utf8]{inputenc}
\usepackage{tabularx}
\usepackage{blindtext}
\usepackage{longtable}
\usepackage{lscape}
\usepackage{setspace}
\usepackage{notoccite} %citation number ordering
\usepackage{lscape} %landscape table
\usepackage{mwe}%%%%%%%%%%%%%%%
\usepackage{booktabs}
\usepackage{amsthm}

\theoremstyle{definition}

\theoremstyle{definition}

\newcommand{\RNum}[1]{\lowercase\expandafter{\romannumeral #1\relax}}
\newcommand{\RNumU}[1]{\uppercase\expandafter{\romannumeral #1\relax}}
\usepackage[numbers]{natbib}

\def\BibTeX{{\rm B\kern-.05em{\sc i\kern-.025em b}\kern-.08em
    T\kern-.1667em\lower.7ex\hbox{E}\kern-.125emX}}
\begin{document}
\title{RVFL-X: A Novel Randomized Network Based on Complex Transformed Real-Valued Tabular Datasets\\
% {\footnotesize \textsuperscript{*}Note: Sub-titles are not captured for https://ieeexplore.ieee.org  and
\thanks{M. Sajid acknowledges the Council of Scientific and Industrial Research (CSIR), New Delhi, for providing fellowship under grants 09/1022(13847)/2022-EMR-I. The code of the paper can be accessed from \url{https://github.com/AnonymousAuthor0011/RVFL-X}.}
}
\author{\IEEEauthorblockN{M. Sajid}
\IEEEauthorblockA{\textit{Department of Mathematics} \\
\textit{Indian Institute of Technology Indore, India}\\
% Indore, India \\
phd2101241003@iiti.ac.in}
\and
\IEEEauthorblockN{Mushir Akhtar}
\IEEEauthorblockA{\textit{Department of Mathematics} \\
\textit{Indian Institute of Technology Indore, India}\\
% Indore, India \\
phd2101241004@iiti.ac.in}
\and
\IEEEauthorblockN{A. Quadir}
\IEEEauthorblockA{\textit{Department of Mathematics} \\
\textit{Indian Institute of Technology Indore, India}\\
% Indore, India \\
mscphd2207141002@iiti.ac.in}
\and
\IEEEauthorblockN{M. Tanveer}
\IEEEauthorblockA{\textit{Department of Mathematics} \\
\textit{Indian Institute of Technology Indore, India}\\
% Indore, India \\
mtanveer@iiti.ac.in}
}
\maketitle
\begin{abstract}
Recent advancements in neural networks, supported by foundational theoretical insights, emphasize the superior representational power of complex numbers. However, their adoption in randomized neural networks (RNNs) has been limited due to the lack of effective methods for transforming real-valued tabular datasets into complex-valued representations. To address this limitation, we propose two methods for generating complex-valued representations from real-valued datasets: a natural transformation and an autoencoder-driven method. Building on these mechanisms, we propose RVFL-X, a complex-valued extension of the random vector functional link (RVFL) network. RVFL-X integrates complex transformations into real-valued datasets while maintaining the simplicity and efficiency of the original RVFL architecture. By leveraging complex components such as input, weights, and activation functions, RVFL-X processes complex representations and produces real-valued outputs. Comprehensive evaluations on $80$ real-valued UCI datasets demonstrate that RVFL-X consistently outperforms both the original RVFL and state-of-the-art (SOTA) RNN variants, showcasing its robustness and effectiveness across diverse application domains.
\end{abstract}

\section{Introduction}
\IEEEPARstart{T}{he} exceptional capability of artificial neural networks (ANNs) to approximate intricate nonlinear mappings has established them as a pivotal tool across a wide spectrum of machine learning applications \cite{ganaie2022ensemble}. ANNs have demonstrated substantial success in diverse domains, including autonomous driving \cite{10817778}, Alzheimer's disease diagnosis \cite{tanveer2024ensemble}, \cite{sajid2024decoding}, \cite{10561527}, Protein structure prediction \cite{senior2020improved} etc.

The backpropagation (BP) algorithm, underpinned by gradient descent (GD), is a widely employed method for optimizing ANN parameters by minimizing the discrepancy between predicted and observed outputs. Despite its widespread use, BP-based optimization is constrained by several limitations, including computational inefficiency, convergence to suboptimal local minima \cite{gori1992problem}, and sensitivity to hyperparameter selection and initialization strategies. These challenges necessitate alternative methodologies for efficient neural network training.

Randomized neural networks (RNNs) \cite{schmidt1992feed} represent a paradigm shift, addressing the drawbacks associated with GD-based optimization. In RNNs, certain parameters are randomly generated and remain fixed during training, and the rest are optimized using closed-form or iterative algorithms \cite{suganthan2021origins}. 

Among the prominent architectures in the RNN, the random vector functional link (RVFL) network \cite{pao1994learning} has gained considerable attention for its simplicity and computational efficiency. The RVFL network, in particular, distinguishes itself through its direct connections between the input and output layers, which contribute to its robust performance. In the RVFL framework, the weights and biases of the hidden layer are initialized randomly and remain fixed throughout the training process. The output parameters, comprising both the direct link weights and those connecting the hidden layer to the output layer, are computed analytically using methods such as the pseudo-inverse or least squares \cite{malik2022random}. The inclusion of direct connections serves as an implicit regularization mechanism, enhancing the network's learning efficacy and robustness \cite{zhang2016comprehensive}. 
% Furthermore, the RVFL's streamlined architecture, relative to ELM, reduces complexity while adhering to theoretical principles such as Occam's razor and the Probably Approximately Correct (PAC) learning framework \cite{kearns1994introduction, shi2021random}. 
The RVFL model also demonstrates rapid training capabilities alongside universal approximation properties \cite{igelnik1995stochastic}.

The RVFL network and its advanced variants have proven effective across various domains, such as handling class imbalance \cite{ganaie2024graph} and noisy datasets \cite{malik2022alzheimer}, particularly in Alzheimer’s disease diagnosis \cite{ganaie2024graph}, \cite{malik2022alzheimer}. These models have become a go-to solution due to their simplicity, efficiency, and versatility in addressing diverse challenges. However, traditional RVFL models, which rely on randomization for feature transformation, often face instability issues during learning. To address these challenges, SP-RVFL \cite{zhang2019unsupervised} introduced a sparse autoencoder with $l1$-norm regularization, enabling more stable and precise learning of network parameters. In parallel, efforts to enhance interpretability led to advanced RVFL models based on granular ball and graph embedding (GB-RVFL and GE-GB-RVFL) \cite{SAJID2025111142}, which improve clarity and transparency of decision making, offering valuable insights into how the models make predictions. Further advancements include the neuro-fuzzy RVFL (NF-RVFL) and neuro-fuzzy ensemble deep RVFL (edRVFL-FIS) models \cite{sajid2024neuro, 10552388}, which integrate human-like reasoning, thus boosting both performance and interpretability. Further, to optimize hidden node selection, IRVFL+ \cite{dai2022incremental} dynamically expands hidden nodes, making the network more adaptable. 
% Meanwhile, KERVFL \cite{chakravorti2020non} incorporates kernel functions into the RVFL framework, effectively eliminating the need for manual hidden node selection and providing greater stability and accuracy. 
Together, these innovations have made RVFL a more powerful and versatile tool in the machine learning landscape.
% This paper advances the RVFL network by enhancing its stability and representational capacity through complex-valued transformations of real-valued tabular data. In the next section, we outline the motivation for this work and present the key contributions.
%%%%%%%%%%%%%%%%
\section{Motivation and Contribution}
The performance of RNN-based models is significantly influenced by the generation of hidden layer features, which play a critical role in the propagation of information to the output layers. In the case of the RVFL network, hidden layer features are generated through random projection. However, this random generation often fails to capture some of the nonlinear features inherent in the data, potentially limiting the model's expressiveness. Prior studies have demonstrated that augmenting RVFL variants with feature enhancement techniques \cite{vukovic2018comprehensive}, \cite{malik2022extended} substantially improves their performance by capturing intricate patterns and relationships in the data. Despite these advancements, the exploration of feature enhancement remains relatively nascent in the context of RVFL-based models.

Recent studies in neural networks, coupled with foundational theoretical analyses, suggest that complex numbers offer a richer representational capacity \cite{danihelka2016associative}, \cite{yadav2023fccns}. They enable robust memory retrieval mechanisms while exhibiting noise-resistant properties. These benefits, however, have been underutilized in RNNs, primarily due to the lack of mechanisms to transform real tabular datasets into complex-valued representations. To address this gap, we propose the RVFL-X model—a novel extension of RVFL that integrates complex transformations into real-valued datasets while maintaining the simplicity and efficiency of the original architecture.

The use of complex parameters introduces several advantages across computational, biological, and signal processing domains. From a computational perspective, \cite{danihelka2016associative} demonstrated the efficiency and numerical stability of holographic reduced representations \cite{plate2003holographic}, which employ complex numbers for associative memory retrieval. Similarly, biologically inspired architectures have leveraged complex weights to enhance the versatility of representations. For example, \cite{reichert2013neuronal} introduced a biologically plausible deep network using complex-valued neurons to model richer and more adaptable representations. These studies emphasize the potential of complex formulations to encode the firing rate and relative timing of neurons.

% From a signal processing standpoint, the phase component of complex numbers holds critical importance. It has been established that phase information in speech signals influences intelligibility \cite{shi2006importance}. Likewise, \citet{oppenheim1981importance} demonstrated that the phase of an image can reconstruct the majority of information encoded in its magnitude, effectively describing shapes, edges, and orientations. This underscores the representational power of phase information for capturing intricate data characteristics.

Despite these theoretical and practical advantages, most existing applications of complex-valued neural networks are restricted to datasets inherently expressed in complex forms, such as those arising from signal processing tasks. The RVFL-X model addresses this limitation by proposing a novel mechanism to transform real tabular datasets into a complex-valued representation. This transformation allows the RVFL-X to retain the architectural simplicity of its real-valued counterpart while exploiting the rich representational capacity of complex-valued features.

To the best of our knowledge, this is the first attempt to systematically integrate complex-valued transformations into real-valued RVFL architectures. The RVFL-X maintains the core properties of the original RVFL model with minimal architectural modifications, enabling the effective capture of nuanced patterns and relationships in real-world datasets. This novel approach paves the way for exploring new avenues in RNNs and unlocking their full potential for a wide range of applications.

The primary highlights of this paper are as follows:  
\begin{itemize}
    \item A novel mechanism is introduced to transform real tabular datasets into complex-valued representations.  
    \item Two transformation techniques are proposed: an intuitive and natural approach; and an autoencoder-based method.
    \item The complex variant of RVFL (RVFL-X) is developed, which converts real-valued tabular data into complex values, employs complex weights and complex activation functions and generates real-valued outputs.
    \item Comprehensive comparisons of RVFL-X with 10 SOTA models are conducted across diverse datasets, including binary and multiclass classification, as well as small and large-scale data.  
\end{itemize}
%%%%%%%%%%%%%%%%%%%
\section{Related Work}
\label{Related_works}
This section establishes the necessary notations and provides an overview of the RVFL network, laying the groundwork for the proposed model.
\vspace{-3mm}
%%%%%%%%%%%%%
\subsection{Notations}
Consider a training dataset represented as $\mathcal{Z} = \{(z_j, w_j) \mid j \in \{1, 2, \ldots, k\}\}$,
where \(z_j \in \mathbb{R}^{1 \times r}\) corresponds to the input features, and \(w_j \in \mathbb{R}^{1 \times d}\) is the associated target vector for \(d\) classes. Here, \(k\) denotes the total number of training samples, and \(r\) indicates the number of attributes. The transpose of a matrix is denoted by \((\cdot)^T\). The matrices representing the inputs and outputs are given by $Z = [z_1^T, z_2^T, \ldots, z_k^T]^T $\text{and} $W = [w_1^T, w_2^T, \ldots, w_k^T]^T$,
respectively.
%%%%%%%%%%%%%%%%%%%%%%%
\subsection{RVFL Neural Network}
The RVFL network, first proposed by \citet{pao1994learning}, is a single-layer feed-forward neural network. It consists of three main components: an input layer, a hidden layer, and an output layer. A distinguishing feature of RVFL is the random initialization of the weights connecting the input to the hidden layer, as well as the biases in the hidden layer. These parameters are fixed and not updated during training. Furthermore, RVFL allows direct connections between the input layer and the output layer, enhancing its ability to model complex relationships.

Let the hidden layer contain $N_h$ neurons. The output of the hidden layer, denoted as $G_1$, is computed as follows:
\begin{align}
    G_1 = \sigma(ZF_w + F_b) \in \mathbb{R}^{k \times N_h},
\end{align}
where $F_w \in \mathbb{R}^{r \times N_h}$ represents the randomly initialized weight matrix with values drawn uniformly from $[-1, 1]$, $F_b \in \mathbb{R}^{1 \times N_h}$ is the bias vector, and $\sigma$ is the activation function.

The output layer combines the contributions from the input features and the hidden layer. Let $G_2$ denote the concatenated matrix of input features $Z$ and hidden layer outputs $G_1$, defined as:
\begin{align}
    G_2 = \begin{bmatrix}
            Z & G_1
          \end{bmatrix}.
\end{align}
The output weights $\eta \in \mathbb{R}^{(r + N_h) \times d}$ are determined by solving the following optimization problem:
\begin{align}
    \eta_{\text{min}} = \underset{\eta}{\arg\min} \, \frac{\mathcal{C}}{2} \|G_2\eta - W\|^2 + \frac{1}{2}\|\eta\|^2,
\end{align}
where $\mathcal{C} > 0$ is a regularization parameter. The closed-form solution for $\eta$ is given by:
\begin{align}
    \eta_{\text{min}} = \begin{cases} 
    {G_2}^T \left(G_2 {G_2}^T + \frac{1}{\mathcal{C}}I\right)^{-1} W, & \text{if } k < (r + N_h), \vspace{2mm} \\
    \left({G_2}^T G_2 + \frac{1}{\mathcal{C}}I\right)^{-1} {G_2}^T W, & \text{if } k \geq (r + N_h),
    \end{cases}
\end{align}
where $I$ is the identity matrix of an appropriate dimension. The term $\hat{W} = G_2\eta_{min}$ IS the predicted output of the network.

\section{Proposed Work}\label{Proposed-work}
Complex-valued neural networks offer significant advantages, however their application has primarily been limited to datasets inherently expressed in complex forms. There is a gap in extending these transformations to real-valued tabular data, limiting the broader potential of neural networks in various domains. This creates an opportunity to develop a more versatile approach that leverages complex-valued representations. We propose RVFL-X, a complex-valued variant of RVFL that transforms real-valued tabular datasets into complex representations. The proposed model preserves the simplicity of the original RVFL while utilizing complex weights and activation functions to capture more intricate patterns and relationships in the data.
\subsection{Conversion of Real-valued Data to Complex-valued Data}\label{sec:3.1}  
This section provides a detailed explanation of the proposed process used to convert real-valued data into complex-valued data. In the subsequent subsection, we apply this methodology to define our proposed RVFL-X model using the complex-converted data. To transform the real-valued tabular dataset into a complex-valued one, we propose two distinct methods: first, an intuitive and natural approach (Section \ref{subsec:3.1.1}), and second, based on an autoencoder-based method (Section \ref{subsec:3.1.2}). 
\subsubsection{\textbf{Natural Method}}\label{subsec:3.1.1} For the numerical dataset $Z$, we define the complex dataset $Z^X$ as follows:
\begin{equation}
Z^X=Z+iS,
\end{equation}
where $S=\mathbf{0}$ is the zero matrix having the same dimension as of $Z$. We call it the natural method because, in the Field of complex numbers, every real number is a complex number, and the natural way to write a real number $z$ into a complex number is $z+i0$, where $0$ is scalar zero.
%%%%%%%%%%%%%%%%%%%%%%%%%%
\subsubsection{\textbf{Autoencoder-Based Method}} \label{subsec:3.1.2}
In this method, the complex dataset \( Z^X \) is constructed from a numerical dataset \( Z \) as:  
\begin{equation}
Z^X = Z + iS,
\end{equation}
where \( S \) is a matrix of the same dimensions as \( Z \), obtained using an autoencoder applied to \( Z \). The autoencoder, consisting of an encoder and a decoder, transforms \( Z \) into a latent representation \( S \), preserving key structural properties. This latent representation serves as the imaginary part of \( Z^X \), enriching the dataset with additional information for improved modeling. This approach leverages the autoencoder's capacity to extract and preserve essential features, ensuring the complex dataset \( Z^X \) is both informative and structurally robust.

The encoder maps the input matrix \( Z \) to the latent representation \( \hat{S} \) through random weights: 
\begin{equation}
\hat{S} = \xi(ZW),
\end{equation}
where \( W \in \mathbb{R}^{r \times r} \) is the weight matrix initialized randomly (e.g., drawn from uniform distributions), and $\xi$ is the normalization function. The decoder reconstructs \( Z \) from \( S \) using 
\begin{equation}
\hat{Z} = \xi(SV),
\end{equation}
where \( V \in \mathbb{R}^{r \times r} \) is the decoder weight matrix. This normalization ensures consistency in scaling during both encoding and decoding while preserving the data’s structural properties.

To learn the latent representation \( S \), the reconstruction error between \( Z \) and \( \hat{Z} \) is minimized through an \( \ell_2 \)-regularized optimization problem, preventing overfitting. The optimization problem is formulated as:
\begin{equation}\label{eq:Autoencoder_Optimization}
V^* = \arg\min_V \frac{\mathcal{C}}{2} \|Z - \hat{S}V\|^2 + \frac{1}{2} \|V\|^2,
\end{equation}
where \( \mathcal{C} \) is a regularization parameter. The closed-form solution for \( V^* \) is given by:
\begin{equation}
V^* = 
\begin{cases} 
\hat{S}^T \left(\hat{S} \hat{S}^T + \frac{1}{\mathcal{C}}I\right)^{-1} Z, & \text{if } k < r, \vspace{2mm} \\
\left(\hat{S}^T \hat{S} + \frac{1}{\mathcal{C}}I\right)^{-1} \hat{S}^T Z, & \text{if } k \geq r,
\end{cases}
\end{equation}
where \( I \) represents the identity matrix. 

Once \( V^* \) is computed, the latent representation \( S \) is obtained from the encoder as:
\begin{equation}
S = \xi\left(Z{\left(\varpi V^*+(1-\varpi){V^*}^T\right)}\right),
\end{equation}
where $\varpi$ takes only two values, i.e., $0$ or $1$, thus $\varpi \in \{0,1\}.$ $\varpi$ is tuned during training. If $\varpi=0$, $V^*$ does not contribute to the formation of the imaginary part $S$, whereas if $\varpi=1$, ${V^*}^T$ is excluded. This latent representation effectively captures critical features of the input data in a transformed space, enhancing robustness and generalization through regularization. Consequently, this representation defines the imaginary part of the data.
%%%%%%%%%%%%%%%%%
\begin{figure}[h]
    \centering
    \includegraphics[width=01\linewidth]{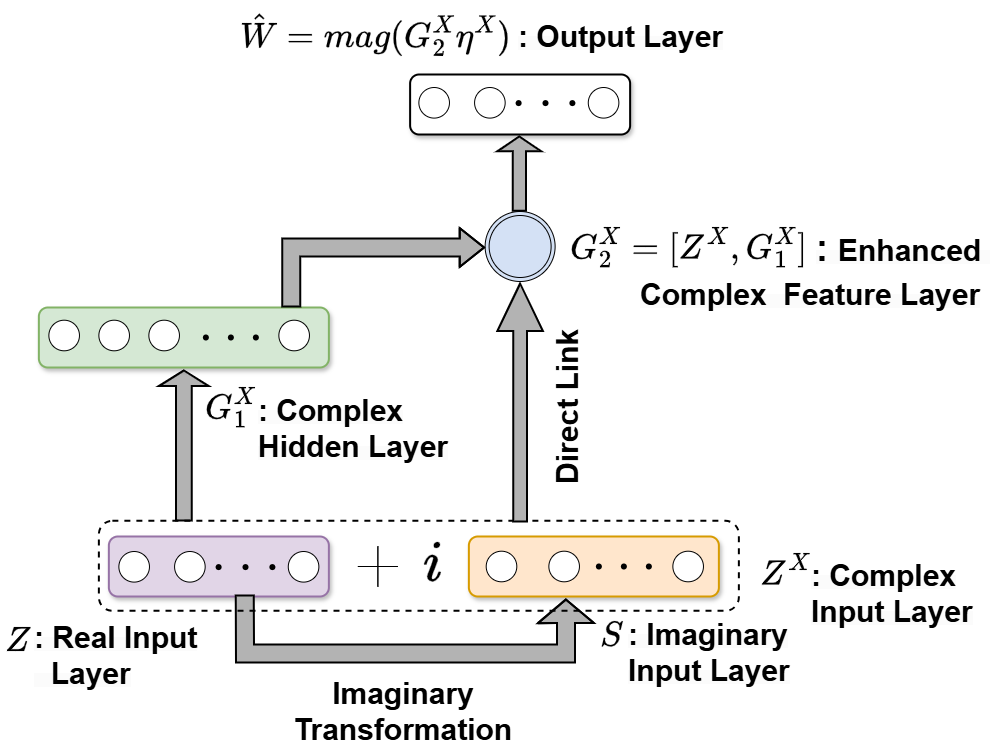}
    \caption{Architecture of the proposed RVFL-X network.}
    \label{fig:RVFL}
\end{figure}
%%%%%%%%%%%%%%%%%%%%%%%%%%%%
\subsection{RVFL-X}\label{Sec:RVFL-X}
The architecture of the proposed RVFL-X network is depicted in Figure~\ref{fig:RVFL}. For the given input matrix $Z$, we convert it to the complex version $Z_X$ and the method of calculating it is given in Section \ref{sec:3.1}. 
Let the complex weight matrix, connecting the input matrix $Z_X$ to the hidden layer and complex bias matrix of the hidden layer be denoted as $F_w^X$ and $F_b^X$, respectively, and defined as follows:
\begin{equation}
    F_w^X = F_w^{real}+iF_w^{imag}
~~\text{and}~~
    F_b^X = F_b^{real}+iF_b^{imag},
\end{equation}
where $F_w^{real} \in \mathbb{R}^{r \times N_h}$ and $F_w^{imag} \in \mathbb{R}^{r \times N_h}$ are generated randomly from the uniform distribution of $[-1,1]$ space and $F_b^{real} \in \mathbb{R}^{r \times N_h}$ and $F_b^{imag} \in \mathbb{R}^{r \times N_h}$ are generated randomly from the uniform distribution of $[0,1]$. All the columns of the $F_b^{real}$ are identical, and similarly for $F_b^{imag}$.

To regularize the model from overfitting, we make some of the entries of the $F_w^X$ and $F_b^X$ to be zero and denoted as $\hat{_{\alpha}F_w^X}$ and $\hat{_{\alpha}F_b^X}$, respectively. The number of entries set to be zero is regulated the hyperparameter $\alpha$, where it denotes the percentage of the entries set to be zero which can be tuned during the model learning process.

The complex hidden layer is generated as follows:
\begin{align}
    G_1^X = \sigma^X(Z^X \cdot_X \hat{_{\alpha}F_w^X} +_X \hat{_{\alpha}F_b^X}) \in \mathbb{R}^{k \times N_h},
\end{align}
where $\cdot_X$ and $+_X$ are complex multiplication and addition, respectively. $\sigma^X$ is modified complex activation function for the real activation function $\sigma$ and defined as follows:
\begin{align}
\sigma^X(z_x) = \sigma(Real(z_x)) + i\sigma(Imag(z_x)),
\end{align}
where $Real(z^X)$ and $Imag(z^X)$ are the real and imaginary part of the complex number $z_x$, respectively.

Next we form the enhanced feature matrix as follows: 
\begin{equation}
G_2^X = [Z^X, G_1^X]
\end{equation}
the predicted output $\hat{W^X}$ of the proposed model can be calculated as: 
\begin{equation}
\hat{W} = mag(G_2^X\eta^X),
\end{equation}
where $\eta^X$ is the complex weight matrix connecting the enhanced feature layer (concatenation of input and hidden layers) to the output layer. $mag(z_x)$ is the magnitude of the complex number $z_x$ and defined as $\sqrt{Real(z_x)^2+Imag(z_x)^2}$. 

The proposed optimization problem can be formulated as:
\begin{align}\label{eq:Complx_Opt_1}
    \eta^X_{\text{min}} = \underset{\eta^X}{\arg\min} \, \frac{\mathcal{C}}{2} \|G_2^X\eta^X - W\|^2 + \frac{1}{2}\|\eta^X\|^2,
\end{align}
where $\mathcal{C} > 0$ is a regularization parameter and same as used in \eqref{eq:Autoencoder_Optimization}. After solving the optimization problem \eqref{eq:Complx_Opt_1} for $\eta^X$, we get:
\begin{align}
    \eta^X_{\text{min}} = \begin{cases} 
    {({G_2}^X)}^T \left({({G_2}^X)} {({G_2}^X)}^T + \frac{1}{\mathcal{C}}I\right)^{-1} W, & \text{if } k < (r + N_h), \vspace{2mm} \\
    \left({({G_2}^X)}^T {({G_2}^X)} + \frac{1}{\mathcal{C}}I\right)^{-1} {({G_2}^X)}^T W, & \text{if } k \geq (r + N_h).
    \end{cases}
\end{align}

As described in Section \ref{sec:3.1}, we proposed two methods for generating the imaginary part of the data. Accordingly, the RVFL-X model based on the method in Section \ref{subsec:3.1.1} will be referred to as RVFL-X-N, while the RVFL-X model based on the method in Section \ref{subsec:3.1.2} will be denoted as RVFL-X-Auto henceforth.
%%%%%%%%%%%%%%%%%%%%%%
\section{Experiments, Results and Discussion}\label{Experiment-section}
\label{experiments}
This section presents comprehensive details of the experimental setup, datasets, and compared models. Subsequently, we delve into the experimental results and conduct statistical analyses. At last, we conduct sensitivity analyses for various hyperparameters of the proposed RVFL-X and ablation study. The code of the paper can be accessed using the following link: \url{https://github.com/AnonymousAuthor0011/RVFL-X}.

\subsection{Compared Models, Datasets and Experimental Setup}
\textbf{Compared Models: }We conduct comparisons among the proposed RVFL-X; and several benchmarks, including RVFL \cite{pao1994learning}, RVFLwoDL (RVFL without direct link, also known as extreme learning machine(ELM)) \cite{huang2006extreme}, intuitionistic fuzzy RVFL (IFRVFL) \cite{malik2022alzheimer}, graph embedded ELM with linear discriminant analysis (GEELM-LDA) \cite{iosifidis2015graph}, graph embedded ELM with local Fisher discriminant analysis (GEELM-LFDA) \cite{iosifidis2015graph}, Total variance RVFL (Total-var-RVFL) \cite{ganaie2020minimum} minimum class variance based ELM (MCVELM) \cite{6542653} and Neuro-fuzzy RVFL-R (NF-RVFL-R), Neuro-fuzzy RVFL-K (NF-RVFL-K), and Neuro-fuzzy RVFL-C (NF-RVFL-C) whose fuzzy layer centers are generated using random-means, K-means and fuzzy C-means \cite{sajid2024neuro}. 

\textbf{Datasets: }To assess the effectiveness of the proposed RVFL-X models, we utilize $80$ benchmark datasets from the UCI repository \cite{dua2017uci}, spanning various domains and sizes. These datasets are categorized into four groups: C-1 category: binary and small, C-2 category: binary and large, C-3 category: multiclass and small, and C-4 category: multiclass and large. Small datasets have sample sizes ranging from 100 to 1,000, while large datasets range from 1,001 to 100,000.

\textbf{Experimental Setup: }The experimental hardware setup includes a personal computer equipped with an Intel(R) Xeon(R) Gold 6226R CPU, operating at a clock speed of 2.90 GHz, and 128 GB of RAM. The system runs Windows 11 and utilizes Matlab 2023a to execute all experiments. Following the experimental protocol in \cite{sajid2024neuro}, we determine the optimal hyperparameters and testing accuracy using grid search along with a five-fold cross-validation technique. Additionally, the hyperparameters for all baseline models are tuned according to the approach outlined in \cite{sajid2024neuro}. For the baseline models, we tune six activation functions: (1) Sigmoid, (2) Sine, (3) Tribas, (4) Radbas, (5) Tansig, and (6) ReLU, as specified in \cite{sajid2024neuro}. For the proposed RVFL-X model, real-valued activation functions are transformed into complex activation functions using the method described in Section \ref{Sec:RVFL-X} of the main manuscript. The regularization parameter $\mathcal{C}$ is chosen from the range $\{10^{-5}, 10^{-4}, \ldots, 10^5\}$. The number of hidden nodes ($N$) is selected from the range of 3 to 203, with a step size of 20. For the proposed model, the percentage parameter $(\alpha)$ is tuned from the set $\{0, 0.1, \ldots, 0.5\}$, where $\alpha = t$ represents the $t \times 100$ percentage that sets the parameters to zero, as discussed in Section \ref{Sec:RVFL-X}.
% \usepackage{graphicx}
% \usepackage[normalem]{ulem}
\begin{table*}[]
\centering
\caption{Average accuracy and rank across 21 C-1 category datasets.}
\label{tab:C1}
\resizebox{\textwidth}{!}{%
\begin{tabular}{lcccccccccccc} \hline \vspace{-3mm}\\ 
\textbf{\textbf{Metric} $\downarrow$ $\mid$ \textbf{Model} $\rightarrow$} &
  \textbf{\textbf{RVFL} \cite{pao1994learning}} &
  \textbf{\textbf{RVFLwoDL} \cite{huang2006extreme}} &
  \textbf{\textbf{IFRVFL} \cite{malik2022alzheimer}} &
  \textbf{\textbf{GEELM-LDA} \cite{iosifidis2015graph}} &
  \textbf{\textbf{GEELM-LFDA} \cite{iosifidis2015graph}} &
  \textbf{\textbf{Total-var-RVFL} \cite{ganaie2020minimum}} &
  \textbf{\textbf{MCVELM} \cite{6542653}} &
  \textbf{\textbf{NF-RVFL-R} \cite{sajid2024neuro}} &
  \textbf{\textbf{NF-RVFL-K} \cite{sajid2024neuro}} &
  \textbf{\textbf{NF-RVFL-C} \cite{sajid2024neuro}} &
  \textbf{\textbf{RVFL-X-N}$^{\dagger}$} &
  \textbf{\textbf{RVFL-X-Auto}$^{\dagger}$} 
  \vspace{1mm}\\ \hline \vspace{-3mm}\\ 
\textbf{Average Accuracy} &
  79.993 &
  79.7806 &
  80.3335 &
  75.703 &
  72.4096 &
  81.1158 &
  80.9615 &
  80.6137 &
  80.884 &
  81.6269 &
  {\ul 82.6226} &
  \textbf{82.8618} 
  \vspace{1mm}\\ \hline \vspace{-3mm}\\ 
\textbf{Average Rank} &
  8.2381 &
  8.9048 &
  6.2381 &
  9.3571 &
  10.5714 &
  5.5 &
  5.5 &
  5.6905 &
  5.5714 &
  5.2619 &
  {\ul 4.5} &
  \textbf{2.6667}
  \vspace{1mm}\\ \hline \vspace{-3mm}\\
\multicolumn{13}{l}{The bold values in each row indicate the best-performing model w.r.t. to the row metric, while the underlined values highlight the second-best-performing model. $\dagger$ denotes the proposed models.}
\end{tabular}%
}
\end{table*}
% \usepackage{graphicx}
% \usepackage[normalem]{ulem}
\begin{table*}[]
\centering
\caption{Average accuracy, rank and standard deviation (Std. Dev.) across 14 C-2 category datasets.}
\label{tab:C2}
\resizebox{\textwidth}{!}{%
\begin{tabular}{lcccccccccccc} \hline \vspace{-3mm}\\ 
\textbf{\textbf{Metric} $\downarrow$ $\mid$ \textbf{Model} $\rightarrow$} &
  \textbf{\textbf{RVFL} \cite{pao1994learning}} &
  \textbf{\textbf{RVFLwoDL} \cite{huang2006extreme}} &
  \textbf{\textbf{IFRVFL} \cite{malik2022alzheimer}} &
  \textbf{\textbf{GEELM-LDA} \cite{iosifidis2015graph}} &
  \textbf{\textbf{GEELM-LFDA} \cite{iosifidis2015graph}} &
  \textbf{\textbf{Total-var-RVFL} \cite{ganaie2020minimum}} &
  \textbf{\textbf{MCVELM} \cite{6542653}} &
  \textbf{\textbf{NF-RVFL-R} \cite{sajid2024neuro}} &
  \textbf{\textbf{NF-RVFL-K} \cite{sajid2024neuro}} &
  \textbf{\textbf{NF-RVFL-C} \cite{sajid2024neuro}} &
  \textbf{\textbf{RVFL-X-N}$^{\dagger}$} &
  \textbf{\textbf{RVFL-X-Auto}$^{\dagger}$} \vspace{1mm}\\ \hline \vspace{-3mm}\\ 
\textbf{Average Accuracy}  & 80.9677 & 80.0556 & 78.3954 & 79.9994 & 76.5356 & 81.5687 & 80.9891 & 80.971 & 81.2448 & 81.6908 & {\ul 82.1302}   & \textbf{83.3793} 
\vspace{1mm}\\ \hline \vspace{-3mm}\\ 
\textbf{Average Rank}      & 7.2857  & 9.5714  & 10.8571 & 8.2143  & 8.1071  & 4.6071  & 6.5     & 7.3214 & 6.5     & 4.1429  & {\ul 2.8214}    & \textbf{2.0714}  
\vspace{1mm}\\ \hline \vspace{-3mm}\\ 
\textbf{Average Std. Dvn.} & 4.4417  & 4.7219  & 5.2407  & 5.3081  & 10.0387 & 4.7577  & 4.4265  & 4.8611 & 4.6659  & 4.9101  & \textbf{4.3711} & {\ul 4.4173}   
\vspace{1mm}\\ \hline \vspace{-3mm}\\
\multicolumn{13}{l}{The bold values in each row indicate the best-performing model w.r.t. to the row metric, while the underlined values highlight the second-best-performing model. $\dagger$ denotes the proposed models.}
\end{tabular}%
}
\end{table*}
%%%%%%%%%%%%%%%
%%%%%%%%%%%%%%%%%
\subsection{Evaluation on C-1 Category Datasets}
Table \ref{tab:C1} summarizes the experimental results, showcasing the average Average accuracy and rank of the proposed RVFL-X models compared to baseline models across 21 C-1 category (binary and small) datasets. A complete accuracy table for each model on each dataset is provided in Table VII of the Appendix. Among them, the proposed RVFL-X-Auto achieves the highest average accuracy of $82.8618\%$, followed by the proposed RVFL-X-N at $82.6226\%$, outperforming other SOTA models. 

Average accuracy can be misleading as it may hide variations in a model's performance across datasets. To address this, we followed \citet{demvsar2006statistical} and conducted statistical tests, including the statistical ranking, Friedman test, and Nemenyi post hoc test. These tests ensured unbiased performance evaluation and enabled broader conclusions about model effectiveness. In the ranking approach, models are ranked based on their dataset-wise performance, with lower ranks for better models. Let $\mathscr{S}$ and $\mathscr{R}$ represent the number of models and datasets, respectively. The rank of the $s^{th}$ model on the $r^{th}$ dataset is $\rho(s,r)$, and its average rank is calculated as $\rho(s,*)=\frac{1}{\mathscr{R}}\sum_{r=1}^{\mathscr{R}}\rho(s,r)$. The last row of Table \ref{tab:C1} shows average ranks, where the proposed RVFL-X-Aut, and RVFL-X models, with ranks $2.6667$ and $4.5$, respectively, outperform baseline models. 

The Friedman test \cite{friedman1937use} compares models' average ranks to identify significant differences. It uses the chi-squared statistic $\chi^2_F$ with $\mathscr{S}-1$ degrees of freedom (dof):  
$\chi^2_F = \frac{12\mathscr{R}}{\mathscr{S}(\mathscr{S}+1)} \left(\sum_{s=1}^{\mathscr{S}} \rho(s,*)^2 - \frac{\mathscr{S}(\mathscr{S}+1)^2}{4}\right).$  
\citet{iman1980approximations} improved $\chi^2_F$ with the $F_F$ statistic:  
$F_F=\chi_F^2 \cdot \frac{(\mathscr{R}-1)}{\mathscr{R}(\mathscr{S}-1)-\chi_F^2}.$ The distribution of $F_F$ is characterized by $(\mathscr{S}-1)$ and $(\mathscr{R}-1)(\mathscr{S}-1)$ dof. 
For $\mathscr{S}=12$ and $\mathscr{R}=21$, $\chi^2_F=92.653$ and $F_F=13.3943$. The $F$-distribution table shows $F_F(11, 220) = 1.8324$ at a $5\%$ significance level. Since $13.3943 > 1.8324$, we reject the null hypothesis, confirming significant differences among the models. 

Further, the Nemenyi post hoc test \cite{demvsar2006statistical} evaluates pairwise statistical differences between models. If the average rank of model $s_1$ is lower than that of $s_2$ by at least the critical difference ($C.D.$), $s_1$ is deemed statistically superior. The $C.D.$ is computed as $C.D. = q_\alpha \sqrt{\frac{\mathscr{S}(\mathscr{S} + 1)}{6\mathscr{R}}},$ where $q_\alpha$ is the critical value from \cite{demvsar2006statistical}. At $\alpha = 0.05$, $C.D. = 3.6363$. Based on the ranks in Table \ref{tab:C1}, both the proposed RVFL-X models are statistically superior to RVFL, RVFLwoDL, GEELM-LDA, and GEELM-LFDA. However, the test fails to establish statistical differences between the proposed models and other baseline methods. Nevertheless, the lowest rank and highest accuracy of the proposed models compared to all existing SOTA models confirm their superior generalization capabilities.
%%%%%%%%%%%%%%%%%
\subsection{Evaluation on C-2 Category Datasets}
Table \ref{tab:C2} presents the experimental results, including the average accuracy, standard deviation, and rank of the proposed RVFL-X models compared to baseline models across 14 C-2 category (binary and large) datasets. Detailed accuracy for each model on individual datasets is provided in Table VIII of the Appendix.

From the results, the RVFL-X-Auto model demonstrates the highest accuracy ($83.3793\%$), the lowest standard deviation ($2.0714$), and the second-lowest rank ($4.4173$). This highlights the model's ability to effectively transform real-valued datasets and process them through the complex architecture of the RVFL network. Notably, the RVFL-X-Auto and RVFL-X-N models achieve approximately $3\%$ higher accuracy than the RVFL and RVFLwoDL networks, further confirming the superior generalization performance of the proposed RVFL-X models. Similarly, the RVFL-X-N model emerges as the second-best in accuracy while achieving the best rank, demonstrating its competitive performance. The low standard deviations of both models reflect the high certainty and stability of their predictions.

Applying the Friedman test yields $F_F = 15.3871$. Referring to the $F$-distribution table, with $F_F(11, 143) = 1.8562$ at a $5\%$ significance level, we reject the null hypothesis, confirming significant differences among the models. Furthermore, the Nemenyi post hoc test calculates a critical difference ($C.D.$) of 4.4535, indicating that both proposed RVFL-X models are statistically superior to RVFL, RVFLwoDL, IF-RVFL, GEELM-LDA, GEELM-LFDA, and NF-RVFL-R. 

Thus the highest accuracy, lowest standard deviation, and superior rankings of the RVFL-X-Auto and RVFL-X-N models demonstrate their exceptional generalization performance. This can be attributed to the proposed RVFL-X models' ability to exploit the rich representational capacity of complex-valued features, which enhances their predictive capabilities compared to existing baseline models.
% \usepackage{graphicx}
% \usepackage[normalem]{ulem}
\begin{table*}[]
\centering
\caption{Average accuracy and rank across 23 C-3 category datasets.}
\label{tab:C3}
\resizebox{\textwidth}{!}{%
\begin{tabular}{lccccccccc} \hline \vspace{-3mm}\\ 
\textbf{\textbf{Metric} $\downarrow$ $\mid$ \textbf{Model} $\rightarrow$} &
  \textbf{\textbf{RVFL} \cite{pao1994learning}} &
  \textbf{\textbf{RVFLwoDL} \cite{huang2006extreme}} &
  \textbf{\textbf{Total-var-RVFL} \cite{ganaie2020minimum}} &
  \textbf{\textbf{MCVELM} \cite{6542653}} &
  \textbf{\textbf{NF-RVFL-R} \cite{sajid2024neuro}} &
  \textbf{\textbf{NF-RVFL-K} \cite{sajid2024neuro}} &
  \textbf{\textbf{NF-RVFL-C} \cite{sajid2024neuro}} &
  \textbf{\textbf{RVFL-X-N}$^{\dagger}$} &
  \textbf{\textbf{RVFL-X-Auto}$^{\dagger}$} 
  \vspace{1mm}\\ \hline \vspace{-3mm}\\ 
\textbf{Average Accuracy} &
  71.385 &
  70.9367 &
  70.9649 &
  71.203 &
  71.6119 &
  72.0101 &
  72.137 &
  {\ul 73.2906} &
  \textbf{74.119} 
  \vspace{1mm}\\ \hline \vspace{-3mm}\\ 
\textbf{Average Rank} &
  6.3478 &
  7.2826 &
  5.2391 &
  5.4783 &
  5.8478 &
  5.1957 &
  4.6522 &
  {\ul 2.6739} &
  \textbf{2.2826}
  \vspace{1mm}\\ \hline \vspace{-3mm}\\
  \multicolumn{10}{l}{The bold values in each row indicate the best-performing model w.r.t. to the row metric, while the underlined values highlight the second-best-performing model. $\dagger$ denotes the proposed models.}
\end{tabular}%
}
\end{table*}
% \usepackage{graphicx}
% \usepackage[normalem]{ulem}
\begin{table*}[]
\centering
\caption{Average accuracy, rank and standard deviation (Std. Dev.) across 22 C-4 category datasets.}
\label{tab:C4}
\resizebox{\textwidth}{!}{%
\begin{tabular}{lccccccccc} \hline \vspace{-3mm}\\ 
\textbf{\textbf{Metric} $\downarrow$ $\mid$ \textbf{Model} $\rightarrow$} &
  \textbf{\textbf{RVFL} \cite{pao1994learning}} &
  \textbf{\textbf{RVFLwoDL} \cite{huang2006extreme}} &
  \textbf{\textbf{Total-var-RVFL} \cite{ganaie2020minimum}} &
  \textbf{\textbf{MCVELM} \cite{6542653}} &
  \textbf{\textbf{NF-RVFL-R} \cite{sajid2024neuro}} &
  \textbf{\textbf{NF-RVFL-K} \cite{sajid2024neuro}} &
  \textbf{\textbf{NF-RVFL-C} \cite{sajid2024neuro}} &
  \textbf{\textbf{RVFL-X-N}$^{\dagger}$} &
  \textbf{\textbf{RVFL-X-Auto}$^{\dagger}$} 
  \vspace{1mm}\\ \hline \vspace{-3mm}\\ 
\textbf{Average Accuracy}  & 85.3395      & 84.5917 & 84.9419 & 84.3676 & 82.7576 & 83.1052 & 85.3942 & \textbf{86.6683} & {\ul 86.6252}   
\vspace{1mm}\\ \hline \vspace{-3mm}\\ 
\textbf{Average Rank}      & 5.2727       & 7.0227  & 4.2727  & 6       & 7.5909  & 6.8409  & 4.5     & {\ul 1.7955}     & \textbf{1.7045} 
\vspace{1mm}\\ \hline \vspace{-3mm}\\ 
\textbf{Average Std. Dvn.} & {\ul 2.8274} & 3.0834  & 4.0949  & 4.1645  & 3.0201  & 3.3523  & 3.1336  & 2.9423           & \textbf{2.8087}
\vspace{1mm}\\ \hline \vspace{-3mm}\\
\multicolumn{10}{l}{The bold values in each row indicate the best-performing model w.r.t. to the row metric, while the underlined values highlight the second-best-performing model. $\dagger$ denotes the proposed models.}
\end{tabular}%
}
\end{table*}
%%%%%%%%%%%%%%%%%%%
\begin{figure}
\begin{minipage}{0.45\linewidth}
\centering
\subfloat[RVFL-X-N]{\label{main:bb}\includegraphics[scale=0.3]{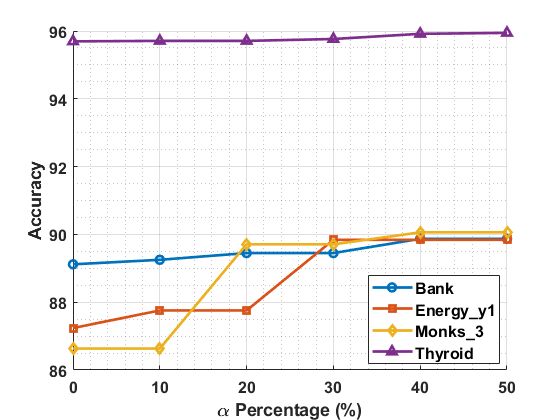}}
\end{minipage}
% \par\medskip
% \par\medskip
\begin{minipage}{0.45\linewidth}
\centering
\subfloat[RVFL-X-Auto]{\label{main:bc}\includegraphics[scale=0.3]{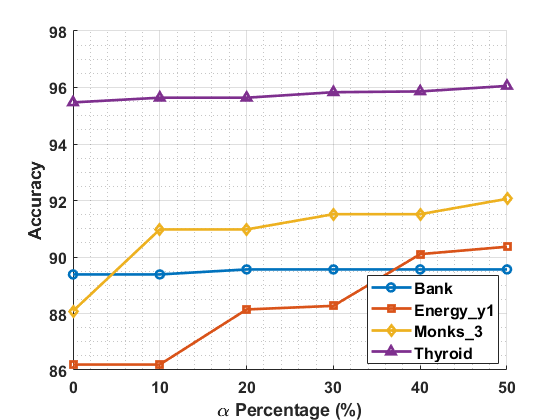}}
\end{minipage}
\par\medskip
\caption{Sensitivity analyses of the proposed RVFL-X-N and RVFL-X-Auto models w.r.t. $\alpha$.}
\label{Fig:sensitivity}
\end{figure}
%%%%%%%%%%%%%%%%%%%%%%%%
\subsection{Evaluation on C-3 Category Datasets}
Table \ref{tab:C3} summarizes the experimental results, presenting the average Average accuracy and rank of the proposed RVFL-X models compared to baseline models across 23 C-3 category (multiclass and small) datasets. Detailed results for each model for every dataset are provided in Table IX of the Appendix.

The RVFL-X-Auto and RVFL-X-N models achieve the highest and second-highest accuracies, $74.119\%$ and $73.2906\%$, respectively, surpassing all baseline models. Notably, the proposed models outperform the NF-RVFL models (a recent SOTA approach leveraging an IF-THEN human-like methodology) by approximately $1\%$. This highlights the remarkable effectiveness of the proposed models. 
% Their simplicity, combined with the ability to efficiently handle complex-valued features, demonstrates both their robustness and adaptability.

In terms of rank, RVFL-X-Auto and RVFL-X-N secure the lowest and second-lowest ranks, $2.2826$ and $2.6739$, respectively, further validating their superior performance. The Friedman test ($F_F = 11.8338 > F_F(11, 143) = 1.9913$ at $\alpha = 0.05$) confirms significant differences among all models on these datasets. Additionally, the Nemenyi post hoc test ($C.D.=2.5051$) reveals that the proposed RVFL-X models are statistically superior to all baseline models except NF-RVFL-C. However, the proposed models achieve lower ranks than NF-RVFL-C, emphasizing their strong generalization capability and consistent performance across diverse datasets.

\subsection{Evaluation on C-4 Category Datasets}
Table \ref{tab:C4} presents the experimental results, including the average accuracy, standard deviation, and rank of the proposed RVFL-X models compared to baseline models across 22 C-4 category (multiclass and large) datasets. Detailed results for each model w.r.t. every dataset are available in Table X of the Appendix.

The proposed RVFL-X-N model achieves the highest accuracy of $86.6683\%$, followed closely by RVFL-X-Auto with $86.6252\%$. In terms of standard deviation and rank, RVFL-X-Auto demonstrates the lowest average standard deviation and rank among all models. The Friedman test ($F_F = 34.2013 > F_F(11, 143) = 1.9939$ at $\alpha = 0.05$) and the Nemenyi post hoc test ($C.D.=2.5614$) further confirm that the proposed models are statistically superior to all baseline models, except Total-Var-RVFL. Nevertheless, the lowest ranks of the proposed models highlight their superior generalization and adaptability.

Notably, the RVFL-X models consistently outperform RVFLwoDL (ELM) in terms of rank and standard deviation, emphasizing the critical role of the direct link in enhancing model performance and stability.
%%%%%%%%%%
\subsection{Sensitivity Analysis}  
Sensitivity analyses of the proposed RVFL-X-N and RVFL-X-Auto models are conducted w.r.t. the percentage parameter $\alpha$, which controls the proportion of zeros in the hidden layer weights and bias matrices for regularization. The results, shown in Fig. \ref{Fig:sensitivity} for four datasets across all categories, illustrate the impact of $\alpha$ values ranging from $0\%$ to $50\%$.  

As $\alpha$ increases, the accuracy of both models improves, demonstrating the effectiveness of regularization through sparsity in the weight and bias matrices. Notably, accuracy stabilizes beyond a certain point, indicating that increasing $\alpha$ beyond this threshold does not yield additional benefits. For instance, when $\alpha = 0.5$, the models achieve near-optimal performance, highlighting this value as a practical choice for maximizing accuracy in the proposed RVFL-X models.
% \usepackage{graphicx}
\begin{table}[]
\centering
\caption{Ablation Study of the RVFL-X-N Model}
\label{tab:Ablation_RVFL-X-N}
\resizebox{9cm}{!}{
\begin{tabular}{lllll}
 \hline \vspace{-3mm}\\ 
Proposed Variant/Dataset           & bank    & energy\_y1 & monks\_3 & thyroid 
\vspace{1mm}\\ \hline \vspace{-3mm}\\
RVFL-X-N                & \textbf{89.8475} & \textbf{91.7927}    & \textbf{92.9615}  & \textbf{96.0417} \\
RVFL-X-N$_{\alpha=0}$     & 89.1175 & 87.2371    & 86.6323  & 95.6944 \\
RVFL-X-N-woDL           & 89.6042 & 89.9711    & 89.8886  & 95.9028 \\
RVFL-X-N-woDL$_{\alpha=0}$ & 89.2502 & 86.0682    & 87.3530  & 95.4722
\vspace{1mm}\\ \hline \vspace{-3mm}\\
\end{tabular}%
}
\end{table}
% \usepackage{graphicx}
\begin{table}[]
\centering
\caption{Ablation Study of the RVFL-X-Auto Model}
\label{tab:Ablation_RVFL-X-Auto}
\resizebox{9cm}{!}{
\begin{tabular}{lllll}
\hline \vspace{-3mm}\\ 
Proposed Variant/Dataset                 & bank    & energy\_y1 & monks\_3 & thyroid 
\vspace{1mm}\\ \hline \vspace{-3mm}\\
RVFL-X-Auto                   & \textbf{89.8032} & \textbf{90.7555}   & \textbf{93.1433} & \textbf{96.0278} \\
RVFL-X-Auto$_{\alpha=0}$      & 89.3829 & 86.1930    & 88.0868 & 95.4722 \\
RVFL-X-Auto-woDL & 89.4050 & 90.4940 & 91.3333 & 95.8750 \\
RVFL-X-Auto-woDL$_{\alpha=0}$ & 89.2723 & 86.4536   & 88.4472 & 95.5000   
\vspace{1mm}\\ \hline \vspace{-3mm}\\
\end{tabular}%
}
\end{table}
%%%%%%%%%%%%%
\subsection{Ablation Study}  
An ablation study is conducted to evaluate the importance of different components in the proposed RVFL-X models. The results are summarized in Table \ref{tab:Ablation_RVFL-X-N} and Table \ref{tab:Ablation_RVFL-X-Auto} for RVFL-X-N and RVFL-X-Auto, respectively. In this study, RVFL-X-N with $\alpha = 0$ (indicating no sparsity) is denoted as RVFL-X-N$_{\alpha=0}$, RVFL-X-N without the direct link is referred to as RVFL-X-N-woDL, and RVFL-X-N without the direct link and with $\alpha = 0$ is labeled as RVFL-X-N-woDL$_{\alpha=0}$. A similar nomenclature is used for RVFL-X-Auto variants.

The results demonstrate that the complete versions of RVFL-X-N and RVFL-X-Auto, which include sparsity controlled by the parameter $\alpha$ and incorporate the direct link, consistently outperform their respective ablated variants. This highlights the critical role of sparsity in weights and biases, as well as the inclusion of the direct link, in achieving superior performance in the proposed models.
%%%%%%%%%%%%%%%%
\section{Conclusion}\label{Conclusions-section}
This paper introduces two innovative mechanisms for transforming real-valued datasets into complex-valued representations: one based on natural transformations and the other leveraging an autoencoder-driven methodology. Using these transformations, we propose RVFL-X, an enhanced extension of the Random Vector Functional Link (RVFL) network, which incorporates complex weights, activation functions, and architectures while producing real-valued outputs.

A rigorous evaluation across 80 UCI datasets, categorized into four groups, demonstrates the superior performance of RVFL-X against 10 SOTA baseline models. Detailed analyses, including accuracy, standard deviation, ranking metrics, and statistical validations such as the Friedman and Nemenyi post hoc tests, affirm the strong generalization and adaptability of RVFL-X. In every evaluation scenario, RVFL-X achieves the highest average accuracy, the lowest standard deviation, and the most favorable rankings, consistently outperforming competing methods. Furthermore, sensitivity analysis and ablation studies highlight the critical role of specific hyperparameters and architectural components in the model’s success. Future extensions of this work could explore deeper and ensemble versions of RVFL-X, with potential applications in domains such as signal processing and beyond.

\bibliographystyle{IEEEtranN}
\bibliography{refs.bib}
% \end{document}
% \clearpage
\newpage
\section{Appendix}

\begin{table*}[htbp]
\centering
\caption{Accuracy of the proposed RVFL-X models and the baseline models on C-1 category datasets.}
\label{tab:Binary_small}
\resizebox{\textwidth}{!}{%
\begin{tabular}{lcccccccccccc} \hline \vspace{-3mm}\\ 
\textbf{\textbf{Dataset} $\downarrow$ $\mid$ \textbf{Model} $\rightarrow$} &
  \textbf{\textbf{RVFL} \cite{pao1994learning}} &
  \textbf{\textbf{RVFLwoDL} \cite{huang2006extreme}} &
  \textbf{\textbf{IFRVFL} \cite{malik2022alzheimer}} &
  \textbf{\textbf{GEELM-LDA} \cite{iosifidis2015graph}} &
  \textbf{\textbf{GEELM-LFDA} \cite{iosifidis2015graph}} &
  \textbf{\textbf{Total-var-RVFL} \cite{ganaie2020minimum}} &
  \textbf{\textbf{MCVELM} \cite{6542653}} &
  \textbf{\textbf{NF-RVFL-R} \cite{10416391}} &
  \textbf{\textbf{NF-RVFL-K} \cite{10416391}} &
  \textbf{\textbf{NF-RVFL-C} \cite{10416391}} &
  \textbf{\textbf{RVFL-X-N}$^{\dagger}$} &
  \textbf{RVFL-X-Auto}$^{\dagger}$    \vspace{1mm}\\ \hline \vspace{-3mm}\\ 
acute\_inflammation          & 100     & 100     & 100     & 100     & 100     & 100     & 100     & 100     & 100     & 100     & 100           & 100              \\
acute\_nephritis             & 100     & 100     & 100     & 95.8333 & 95.8333 & 100     & 100     & 100     & 100     & 100     & 100           & 100              \\
breast\_cancer\_wisc         & 88.5653 & 87.9897 & 89.8499 & 86.5591 & 84.2724 & 88.704  & 88.8448 & 88.2785 & 88.4183 & 87.705  & 90.2785       & 89.9897          \\
chess\_krvkp                 & 72.0313 & 71.936  & 72.6262 & 76.7814 & 69.2351 & 73.7833 & 73.7828 & 82.6045 & 81.7908 & 85.2322 & 83.0741       & 84.5757          \\
congressional\_voting        & 63.6782 & 63.2184 & 58.8506 & 59.7701 & 54.2529 & 63.908  & 63.4483 & 63.908  & 63.908  & 64.1379 & 63.6782       & 64.8276          \\
credit\_approval             & 85.2174 & 85.3623 & 86.5217 & 85.5072 & 85.6522 & 85.5072 & 85.5072 & 85.942  & 85.942  & 85.7971 & 85.5072       & 86.087           \\
cylinder\_bands              & 66.4154 & 65.8081 & 63.4875 & 67.5405 & 67.5385 & 67.7727 & 66.0156 & 68.9606 & 69.724  & 70.1294 & 70.1256       & 70.3103          \\
echocardiogram               & 84.6724 & 83.9031 & 80.7692 & 73.2194 & 68.661  & 85.4701 & 85.4986 & 85.4416 & 85.4416 & 85.4416 & 85.4416       & 86.2108          \\
haberman\_survival           & 73.4902 & 73.8181 & 75.1348 & 55.5632 & 52.2898 & 74.4738 & 73.4902 & 75.1296 & 75.1296 & 75.4574 & 74.146        & 75.4574          \\
ilpd\_indian\_liver          & 71.5311 & 72.2149 & 72.7277 & 61.7521 & 61.4353 & 72.5626 & 72.5508 & 72.3932 & 72.3917 & 72.2178 & 72.3814       & 73.4188          \\
mammographic                 & 79.9196 & 79.2978 & 79.7107 & 75.7572 & 73.7932 & 80.0237 & 80.2332 & 79.2957 & 79.504  & 79.1899 & 80.2321       & 79.1942          \\
monks\_1                     & 83.7934 & 82.1654 & 84.6895 & 73.5811 & 71.5476 & 84.1538 & 84.8697 & 81.4527 & 82.3584 & 84.342  & 93.5151       & 90.8156          \\
monks\_2                     & 80.0138 & 81.6804 & 82.6722 & 60.7273 & 57.23   & 85.8485 & 82.8499 & 85.8485 & 87.3774 & 87.3774 & 89.0041       & 87.5028          \\
monks\_3                     & 91.1564 & 90.7912 & 90.0704 & 82.8419 & 74.5602 & 91.6986 & 91.8821 & 91.8436 & 91.6486 & 92.1744 & 92.9615       & 93.1433          \\
pima                         & 72.0075 & 72.2698 & 73.8282 & 66.401  & 54.0625 & 72.92   & 72.9166 & 73.6983 & 72.9174 & 72.6594 & 73.824        & 74.3545          \\
pittsburg\_bridges\_T\_OR\_D & 87.1905 & 87.1905 & 89.1905 & 82.1429 & 80.1905 & 89.1429 & 90.1429 & 90.1905 & 88.2381 & 90.1905 & 90.1429       & 90.2381          \\
planning                     & 71.3814 & 71.3814 & 69.8048 & 59.1592 & 61.1261 & 72.4925 & 73.048  & 73.018  & 71.9369 & 71.9369 & 72.4925       & 73.033           \\
spect                        & 68.3019 & 66.7925 & 68.3019 & 63.3962 & 62.6415 & 69.0566 & 67.9245 & 69.0566 & 69.434  & 68.6792 & 68.3019       & 70.1887          \\
statlog\_heart               & 80.3704 & 80      & 81.8519 & 73.7037 & 74.8148 & 81.4815 & 82.5926 & 81.4815 & 82.2222 & 81.8519 & 81.8519       & 82.2222          \\
tic\_tac\_toe                & 88.8264 & 88.9278 & 81.4316 & 94.6875 & 79.2059 & 91.8521 & 90.0769 & 65.3125 & 68.5684 & 79.3216 & 94.5681       & 84.6564          \\
vertebral\_column\_2clases   & 71.2903 & 70.6452 & 85.4839 & 94.8387 & 92.2581 & 72.5806 & 74.5161 & 79.0323 & 81.6129 & 80.3226 & 73.5484       & 83.871 \vspace{1mm}\\ \hline \vspace{-3mm}\\ 
\textbf{Average}             & 79.993  & 79.7806 & 80.3335 & 75.703  & 72.4096 & 81.1158 & 80.9615 & 80.6137 & 80.884  & 81.6269 & {\ul 82.6226} & \textbf{82.8618}
\vspace{1mm}\\ \hline \vspace{-3mm}\\ 
\end{tabular}%
}
\end{table*}
% \usepackage{graphicx}
% \usepackage[normalem]{ulem}
\begin{table*}[]
\centering
\caption{Accuracy of the proposed RVFL-X models and the baseline models on C-2 category datasets.}
\label{tab:binary_large}
\resizebox{\textwidth}{!}{%
\begin{tabular}{lcccccccccccc} \hline \vspace{-3mm}\\ 
\textbf{\textbf{Dataset} $\downarrow$ $\mid$ \textbf{Model} $\rightarrow$} &
  \textbf{\textbf{RVFL} \cite{pao1994learning}} &
  \textbf{\textbf{RVFLwoDL} \cite{huang2006extreme}} &
  \textbf{\textbf{IFRVFL} \cite{malik2022alzheimer}} &
  \textbf{\textbf{GEELM-LDA} \cite{iosifidis2015graph}} &
  \textbf{\textbf{GEELM-LFDA} \cite{iosifidis2015graph}} &
  \textbf{\textbf{Total-var-RVFL} \cite{ganaie2020minimum}} &
  \textbf{\textbf{MCVELM} \cite{6542653}} &
  \textbf{\textbf{NF-RVFL-R} \cite{10416391}} &
  \textbf{\textbf{NF-RVFL-K} \cite{10416391}} &
  \textbf{\textbf{NF-RVFL-C} \cite{10416391}} &
  \textbf{\textbf{RVFL-X-N}$^{\dagger}$} &
  \textbf{RVFL-X-Auto}$^{\dagger}$    \vspace{1mm}\\ \hline \vspace{-3mm}\\ 
adult                             & 84.0342 & 83.8725 & 83.5438 & 83.422  & 83.1784 & 84.0178 & 83.9011 & 83.2992 & 83.5879 & 84.6218 & 84.6976       & 84.6259          \\
bank                              & 89.4051 & 89.4934 & 89.1173 & 89.6042 & 46.4966 & 89.6705 & 89.6703 & 89.4269 & 89.4492 & 89.9139 & 89.8475       & 89.8032          \\
breast\_cancer\_wisc\_diag        & 93.8503 & 92.2683 & 89.2719 & 93.1455 & 86.988  & 94.1997 & 93.4995 & 94.3751 & 94.3782 & 95.6063 & 95.4324       & 95.08            \\
breast\_cancer\_wisc\_prog        & 81.359  & 80.3846 & 78.359  & 62.0897 & 62.8462 & 82.359  & 81.8718 & 83.359  & 81.8846 & 80.9103 & 81.8718       & 82.8462          \\
conn\_bench\_sonar\_mines\_rocks  & 62.079  & 60.5226 & 54.8316 & 80.5343 & 73.6585 & 64.4251 & 64.4251 & 65.8885 & 66.2602 & 63.4611 & 63.4959       & 74.007           \\
connect\_4                        & 75.4518 & 75.4059 & 75.3407 & 76.7316 & 75.4281 & 75.4992 & 75.4459 & 75.4844 & 75.5022 & 76.2156 & 77.2266       & 77.0017          \\
hill\_valley                      & 82.2603 & 79.7014 & 80.2772 & 76.9231 & 80.3846 & 83.6592 & 80.0289 & 82.8429 & 82.6732 & 83.0874 & 83.0028       & 85.9742          \\
ionosphere                        & 88.6358 & 86.9175 & 84.3581 & 83.2274 & 82.664  & 89.4809 & 88.326  & 88.3461 & 89.7626 & 88.9095 & 91.7505       & 90.8974          \\
magic                             & 78.7487 & 78.5279 & 77.5289 & 77.4289 & 78.3579 & 78.7171 & 78.7592 & 76.3407 & 76.3302 & 78.4017 & 79.3218       & 82.224           \\
mushroom                          & 96.3931 & 96.0735 & 93.9684 & 96.3071 & 98.7561 & 97.2919 & 96.7382 & 92.6761 & 95.9132 & 99.3351 & 97.7596       & 97.6118          \\
oocytes\_merluccius\_nucleus\_4d  & 82.2884 & 81.5045 & 79.1554 & 81.9957 & 82.7273 & 82.3883 & 82.1942 & 81.6021 & 81.7006 & 82.1923 & 83.5643       & 83.4605          \\
oocytes\_trisopterus\_nucleus\_2f & 78.9419 & 77.5182 & 75.2219 & 78.0694 & 79.8889 & 79.6013 & 79.6025 & 79.3863 & 79.3779 & 79.9345 & 80.8119       & 82.4566          \\
ringnorm                          & 51.5541 & 51.5405 & 51.473  & 51.5541 & 51.5784 & 52.0405 & 51.9459 & 51.6081 & 51.6486 & 51.6892 & 51.7973       & 52.0135          \\
spambase                          & 88.546  & 87.0472 & 85.0883 & 88.9582 & 88.546  & 88.6116 & 87.4382 & 88.9582 & 88.9583 & 89.3931 & 89.2423       & 89.3075          \vspace{1mm}\\ \hline \vspace{-3mm}\\ 
\textbf{Average}                  & 80.9677 & 80.0556 & 78.3954 & 79.9994 & 76.5356 & 81.5687 & 80.9891 & 80.971  & 81.2448 & 81.6908 & {\ul 82.1302} & \textbf{83.3793}
\vspace{1mm}\\ \hline \vspace{-3mm}\\ 
\end{tabular}%
}
\end{table*}
% \usepackage{graphicx}
% \usepackage[normalem]{ulem}
\begin{table*}[]
\centering
\caption{Accuracy of the proposed RVFL-X models and the baseline models on C-3 category datasets.}
\label{tab:multi_small}
\resizebox{\textwidth}{!}{%
\begin{tabular}{lccccccccc} \hline \vspace{-3mm}\\ 
\textbf{\textbf{Dataset} $\downarrow$ $\mid$ \textbf{Model} $\rightarrow$} &
  \textbf{\textbf{RVFL} \cite{pao1994learning}} &
  \textbf{\textbf{RVFLwoDL} \cite{huang2006extreme}} &
  \textbf{\textbf{Total-var-RVFL} \cite{ganaie2020minimum}} &
  \textbf{\textbf{MCVELM} \cite{6542653}} &
  \textbf{\textbf{NF-RVFL-R} \cite{10416391}} &
  \textbf{\textbf{NF-RVFL-K} \cite{10416391}} &
  \textbf{\textbf{NF-RVFL-C} \cite{10416391}} &
  \textbf{\textbf{RVFL-X-N}$^{\dagger}$} &
  \textbf{RVFL-X-Auto}$^{\dagger}$    \vspace{1mm}\\ \hline \vspace{-3mm}\\ 
annealing                  & 89.6381 & 88.1918 & 89.1937 & 88.6381 & 86.6313 & 87.0782 & 90.8591 & 91.0832       & 91.4165         \\
audiology\_std             & 69.3205 & 65.7564 & 50.3718 & 51.8846 & 71.3846 & 70.3718 & 67.8077 & 71.359        & 70.3462         \\
balance\_scale             & 98.4    & 98.4    & 98.56   & 98.56   & 98.4    & 98.24   & 98.24   & 98.72         & 98.56           \\
car                        & 72.6265 & 72.1057 & 72.6234 & 72.2758 & 70.0139 & 70.8337 & 71.0626 & 73.2048       & 72.1642         \\
contrac                    & 40.458  & 40.1183 & 41.3386 & 41.0688 & 41.4934 & 42.3673 & 41.5416 & 41.6089       & 45.6813         \\
dermatology                & 97.5379 & 97.0011 & 97.2714 & 97.5417 & 97.5454 & 97.8193 & 97.5379 & 97.8119       & 97.5417         \\
ecoli                      & 60.9175 & 61.2116 & 51.0667 & 51.0667 & 60.619  & 60.6277 & 61.5145 & 60.9175       & 62.7085         \\
energy\_y1                 & 88.6716 & 88.2777 & 89.7106 & 89.0604 & 87.889  & 88.5434 & 89.0595 & 91.7927       & 90.7555         \\
energy\_y2                 & 90.4932 & 90.4932 & 91.14   & 90.7538 & 89.3269 & 89.1928 & 89.8404 & 91.9243       & 92.0567         \\
glass                      & 37.6855 & 39.0808 & 34.4297 & 37.6855 & 42.3477 & 42.8239 & 40.4873 & 39.546        & 44.1971         \\
hayes\_roth                & 61.875  & 60.625  & 62.5    & 61.875  & 61.875  & 61.875  & 65      & 65            & 65              \\
heart\_cleveland           & 59.7268 & 59.377  & 61.0328 & 59.3607 & 61.0383 & 61.3388 & 60.0437 & 61.0383       & 61.6721         \\
heart\_va                  & 40      & 40      & 40.5    & 40      & 41      & 42.5    & 41.5    & 41            & 41.5            \\
iris                       & 74.6667 & 74      & 74.6667 & 77.3333 & 73.3333 & 74.6667 & 74.6667 & 79.3333       & 85.3333         \\
led\_display               & 72.6    & 72.5    & 73.5    & 72.8    & 72.7    & 72.2    & 73.4    & 73.6          & 73.5            \\
lymphography               & 86.4138 & 85.0345 & 86.4138 & 86.4368 & 85.7701 & 87.1034 & 88.5057 & 88.4368       & 87.7931         \\
nursery                    & 70.3935 & 70.1775 & 63.0093 & 62.8627 & 66.7824 & 66.3889 & 71.9444 & 78.1481       & 77.1373         \\
pittsburg\_bridges\_REL\_L & 62.1429 & 63.1429 & 64.1429 & 67.0952 & 67.2381 & 69.0476 & 62.1905 & 65.1905       & 68.0952         \\
seeds                      & 89.0476 & 87.1429 & 90      & 89.0476 & 90.4762 & 90      & 90.4762 & 90            & 90.4762         \\
soybean                    & 87.6954 & 86.9579 & 88.4264 & 88.275  & 87.3991 & 88.1387 & 90.4723 & 90.3231       & 90.1803         \\
teaching                   & 69.3763 & 69.3763 & 70.6882 & 72.7097 & 70.7312 & 70.043  & 70.7097 & 72.0215       & 72.6882         \\
vertebral\_column\_3clases & 65.1613 & 65.1613 & 84.1935 & 84.1935 & 67.4194 & 69.0323 & 65.4839 & 65.8065       & 68.3871         \\
yeast                      & 57.0077 & 57.4117 & 57.4133 & 57.1444 & 55.6593 & 56.0005 & 56.8064 & 57.8171       & 57.5475        
\vspace{1mm}\\ \hline \vspace{-3mm}\\ 
\textbf{Average}           & 71.385  & 70.9367 & 70.9649 & 71.203  & 71.6119 & 72.0101 & 72.137  & {\ul 73.2906} & \textbf{74.119}
\vspace{1mm}\\ \hline \vspace{-3mm}\\ 
\end{tabular}%
}
\end{table*}
% \usepackage{graphicx}
% \usepackage[normalem]{ulem}
\begin{table*}[]
\centering
\caption{Accuracy of the proposed RVFL-X models and the baseline models on C-4 category datasets.}
\label{tab:multi_large}
\resizebox{\textwidth}{!}{%
\begin{tabular}{lccccccccc} \hline \vspace{-3mm}\\ 
\textbf{\textbf{Dataset} $\downarrow$ $\mid$ \textbf{Model} $\rightarrow$} &
  \textbf{\textbf{RVFL} \cite{pao1994learning}} &
  \textbf{\textbf{RVFLwoDL} \cite{huang2006extreme}} &
  \textbf{\textbf{Total-var-RVFL} \cite{ganaie2020minimum}} &
  \textbf{\textbf{MCVELM} \cite{6542653}} &
  \textbf{\textbf{NF-RVFL-R} \cite{10416391}} &
  \textbf{\textbf{NF-RVFL-K} \cite{10416391}} &
  \textbf{\textbf{NF-RVFL-C} \cite{10416391}} &
  \textbf{\textbf{RVFL-X-N}$^{\dagger}$} &
  \textbf{RVFL-X-Auto}$^{\dagger}$    \vspace{1mm}\\ \hline \vspace{-3mm}\\ 
abalone                          & 63.4665 & 63.4419 & 63.754  & 63.6578 & 63.8253 & 64.1365 & 64.0887 & 63.8735          & 64.1126       \\
cardiotocography\_10clases       & 65.9924 & 65.2894 & 66.1351 & 66.2307 & 62.9851 & 62.8425 & 69.2395 & 71.6383          & 71.3568       \\
cardiotocography\_3clases        & 86.1733 & 85.5622 & 86.0808 & 85.9382 & 84.6213 & 84.2003 & 85.4674 & 86.3151          & 86.2216       \\
conn\_bench\_vowel\_deterding    & 95.8586 & 95.1515 & 96.1616 & 95.7576 & 87.9798 & 88.7879 & 92.2222 & 96.1616          & 96.3636       \\
image\_segmentation              & 87.6623 & 87.0996 & 88.3117 & 87.4459 & 83.2035 & 83.5931 & 83.6797 & 89.4372          & 89.2641       \\
letter                           & 80.505  & 79.695  & 80.67   & 79.86   & 61.14   & 63.205  & 81.605  & 82.21            & 82.33         \\
low\_res\_spect                  & 87.7605 & 86.6285 & 73.9852 & 74.1721 & 87.3849 & 87.5666 & 87.7605 & 88.8873          & 88.8873       \\
oocytes\_merluccius\_states\_2f  & 91.2846 & 90.8948 & 91.8737 & 91.483  & 91.3845 & 91.1894 & 92.067  & 92.3625          & 92.4605       \\
oocytes\_trisopterus\_states\_5b & 87.9385 & 85.853  & 89.3647 & 87.8286 & 87.7205 & 87.2816 & 87.7175 & 89.2584          & 88.9221       \\
optical                          & 95.4626 & 94.3594 & 95.4626 & 94.8754 & 93.3808 & 94.0214 & 97.0107 & 97.4733          & 97.4021       \\
page\_blocks                     & 95.4137 & 95.4137 & 95.4136 & 95.3952 & 94.6645 & 94.7012 & 94.7558 & 95.5963          & 95.5416       \\
pendigits                        & 98.5353 & 98.4716 & 98.6354 & 98.5262 & 92.7402 & 94.6233 & 97.5345 & 98.6262          & 98.6718       \\
semeion                          & 87.0673 & 82.4233 & 87.2562 & 83.0492 & 88.0726 & 88.5732 & 87.3187 & 90.2072          & 89.8925       \\
statlog\_image                   & 95.1948 & 94.8918 & 95.6277 & 95.2814 & 91.5584 & 91.5584 & 93.0736 & 96.0606          & 96.1905       \\
statlog\_landsat                 & 82.0513 & 81.4141 & 82.2999 & 81.7249 & 80.4196 & 81.3209 & 81.2898 & 82.6107          & 82.6884       \\
statlog\_shuttle                 & 98.6483 & 98.6241 & 98.681  & 98.6793 & 98.6345 & 98.6638 & 98.6724 & 98.7224          & 98.7328       \\
statlog\_vehicle                 & 82.0327 & 81.5593 & 82.0313 & 81.3227 & 79.1974 & 79.5475 & 82.2666 & 83.2175          & 83.3317       \\
thyroid                          & 95.625  & 95.4722 & 95.7917 & 95.5833 & 94.6667 & 94.7639 & 96.0278 & 96.0417          & 96.0278       \\
wall\_following                  & 75.6057 & 75.239  & 75.734  & 75.1845 & 71.518  & 72.1223 & 79.583  & 81.1777          & 80.2244       \\
waveform                         & 86.54   & 85.98   & 86.56   & 86.26   & 86.88   & 86.68   & 87.22   & 87.26            & 87.24         \\
waveform\_noise                  & 86.28   & 85.02   & 86.42   & 85.42   & 86.28   & 86.3    & 86.58   & 86.46            & 86.46         \\
wine\_quality\_white             & 52.3699 & 52.5332 & 52.472  & 52.4103 & 52.4106 & 52.6349 & 53.4925 & 53.1047          & 53.4311       \vspace{1mm}\\ \hline \vspace{-3mm}\\ 
\textbf{Average}                 & 85.3395 & 84.5917 & 84.9419 & 84.3676 & 82.7576 & 83.1052 & 85.3942 & \textbf{86.6683} & {\ul 86.6252}
\vspace{1mm}\\ \hline \vspace{-3mm}\\ 
\end{tabular}%
}
\end{table*}
% \newpage
% \bibliographystyle{IEEEtranN}
% \bibliography{refs.bib}
\end{document}